\documentclass[10pt,twocolumn,letterpaper]{article}

\usepackage[21]{cvpr} 

\definecolor{cvprblue}{rgb}{0.21,0.49,0.74}
\usepackage[pagebackref,breaklinks,colorlinks,allcolors=cvprblue]{hyperref}
\usepackage{indentfirst}
\usepackage{colortbl}
\usepackage{times}
\usepackage{latexsym}
\usepackage{tabularx}
\usepackage{booktabs}
\usepackage{amsmath}
\usepackage{float}
\usepackage{multirow}
\usepackage{graphicx}
\usepackage{makecell,array}
\usepackage{outlines}
\usepackage[accsupp]{axessibility}  

\def\paperID{   } 
\def\confName{CVPR}
\def\confYear{2026}

\title{Revisiting Multimodal KV Cache Compression: \\ A Frequency-Domain-Guided Outlier-KV-Aware Approach}

\author{Yaoxin Yang\\
\and
Peng Ye\\
\and 
Xudong Tan\\
\and
Chongjun Tu\\
\and
Maosen Zhao\\
\and
Jia Hao\\
\and
Tao Chen\\
}
\author{Yaoxin Yang$^{1,5}$\footnotemark[1] \quad
Peng Ye$^{3,4}$\footnotemark[1] \quad
Xudong Tan$^1$ \quad
Chongjun Tu$^1$ \quad
Maosen Zhao$^1$ \quad
\\
Jia Hao$^5$ \quad
Tao Chen$^{2,1}$\footnotemark[2]\\
$^1$College of Future Information Technology, Fudan University\\
$^2$Shanghai Innovation Institute. \\ 
$^3$The Chinese University of Hong Kong \\
$^4$Shanghai Artificial Intelligence Laboratory \\
$^5$Zhangjiang Laboratory \\
{\tt\small yxyang24@m.fudan.edu.cn eetchen@fudan.edu.cn}
}
\begin{document}
\maketitle
\renewcommand{\thefootnote}{\fnsymbol{footnote}}
\footnotetext[2]{Corresponding author. \footnotemark[1]Equal contribution.}
\vspace{-5mm}
\begin{abstract}

Multimodal large language models 
suffer from substantial inference overhead since multimodal KV Cache grows proportionally with the visual input length. Existing multimodal KV Cache compression methods mostly rely on attention score to reduce cache size, which makes them are incompatible with established efficient attention kernels (e.g., FlashAttention) and ignores the contribution of value vectors to the attention output.
In this work, we revisit multimodal KV Cache compression from the perspective of the KV matrices’ distribution. First, we observe that frequency-domain energy of multimodal KV matrices is predominantly concentrated in low-frequency and extract this principal energy via a low-pass filter. 
Further, we find that removing KV pairs that deviate substantially from this principal energy leads to a pronounced performance drop, which we define as \textbf{Outlier KVs}.  
Considering Outlier KVs are more likely to encode features critical for inference, we propose \textbf{FlashCache}, a frequency-domain–guided, Outlier-KV-aware KV Cache compression framework. First, we introduce an Outlier KV Recognition Module that models the principal component of multimodal KV matrices in the frequency domain and preferentially retains KV pairs that significantly deviate from it. Furthermore, Dynamic Budget Allocation Module is designed to adaptively determine the per-layer KV Cache size to retain more Outlier KVs. Experiments on multiple MLLMs and benchmarks demonstrate that FlashCache outperforms state-of-the-art multimoal KV compression methods, achieving up to 1.69× faster decoding with 80\% lower KV memory usage while maintaining task performance.

\end{abstract}

\section{Introduction}
\label{sec:intro}
\begin{figure}
    \centering
    \includegraphics[width=\linewidth]{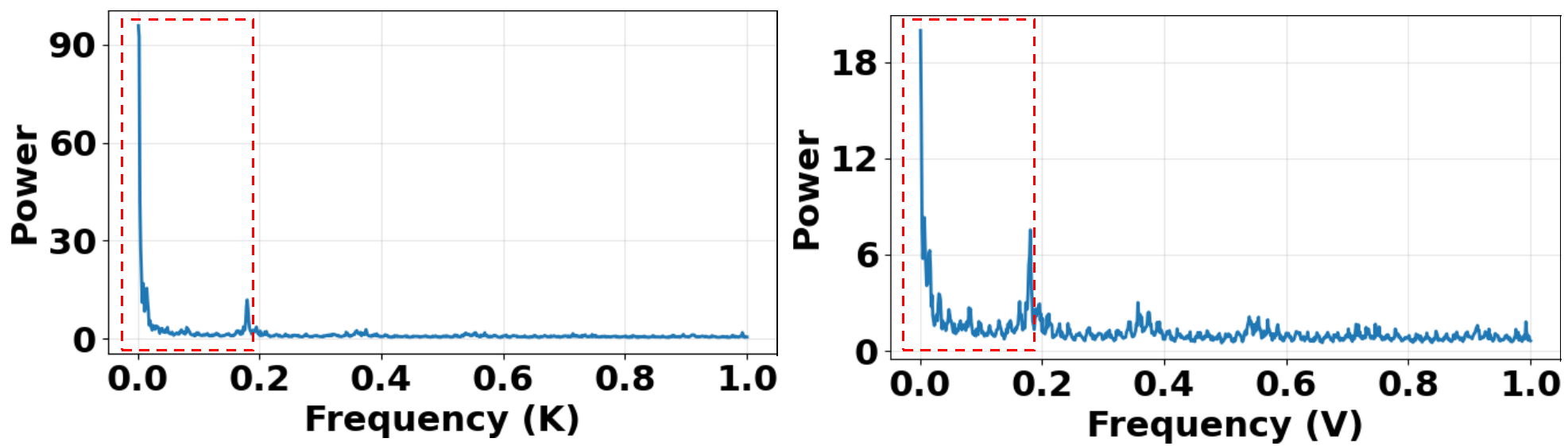}
    \caption{Frequency-domain energy distribution of the KV matrices. We observe that the frequency-domain energy of KV matrices is predominantly concentrated at low frequency, with high frequency components occupying a relatively small proportion.
    The red dashed box in the figure highlights the low-frequency concentration phenomenon. Experiments are conducted on MileBench~\cite{song2024milebench} with Qwen2.5-VL-7B.}
    \label{fig1}
\vspace{-5mm}
\end{figure}
Multimodal large language models (MLLMs)~\cite{achiam2023gpt,zhu2024multilingual,chen2024internvl,li2024survey,liu2024improved,wang2024qwen2} exhibit strong multimodal perception and reasoning, yet their practical deployment is hindered by high computation and memory costs, driven primarily by the substantial volume of visual tokens at inference time~\cite{jin2024efficient,huang2024dynamic,ye2025beta}.
To address this, many techniques, such as token pruning~\cite{chen2024image,he2025zipvl,tan2025tokencarve,zhang2024sparsevlm}, model quantization~\cite{yu2025mquant,xie2024advancing}, and KV Cache optimizations~\cite{adnan2024keyformer,hooper2024kvquant,li2024survey1}, have been proposed to accelerate MLLM inference. Among them, KV Cache improves decoding efficiency by storing the Keys and Values of previously seen tokens,  avoiding repeated prefix computation during decoding, thereby garnering significant attention~\cite{shi2024keep,ge2023model}. Despite achieving impressive progress, as context length grows (e.g., multi-image, high-resolution and video settings~\cite{song2024milebench,meng2024mmiu,hrbench,tu2025favor}), the length of multimodal KV Cache increases accordingly, leading to substantial GPU memory overhead and a sharp slowdown during the decoding stage.

Recently, several works have attempted to prune redundant portions of long-context KV Cache in MLLMs to further reduce memory usage and speed up decoding. LOOK-M~\cite{wan2024look} observes extensive redundancy in visual KVs within long multimodal contexts and compresses the KV Cache by screening and merging visual KV pairs with low attention score during the prefill stage. MEDA~\cite{wan2025meda} uses cross-modal attention entropy to allocate an appropriate KV Cache size per MLLM layer. However, these approaches all rely on attention score to shrink the KV Cache, which raises several issues: (1) efficient attention kernels (e.g., FlashAttention~\cite{dao2022flashattention}) typically do not explicitly output the full attention score, and recomputing attention score introduces additional overhead; (2) since attention score are obtained via Query–Key dot products, directly compressing KV Cache via attention score ignores the information contribution of the Value matrix~\cite{senguptavalue}. This naturally leads to the question: \textit{how can we perform efficient multimodal KV Cache compression without relying on attention score}?

\begin{figure}
    \centering
    \includegraphics[width=0.9\linewidth]{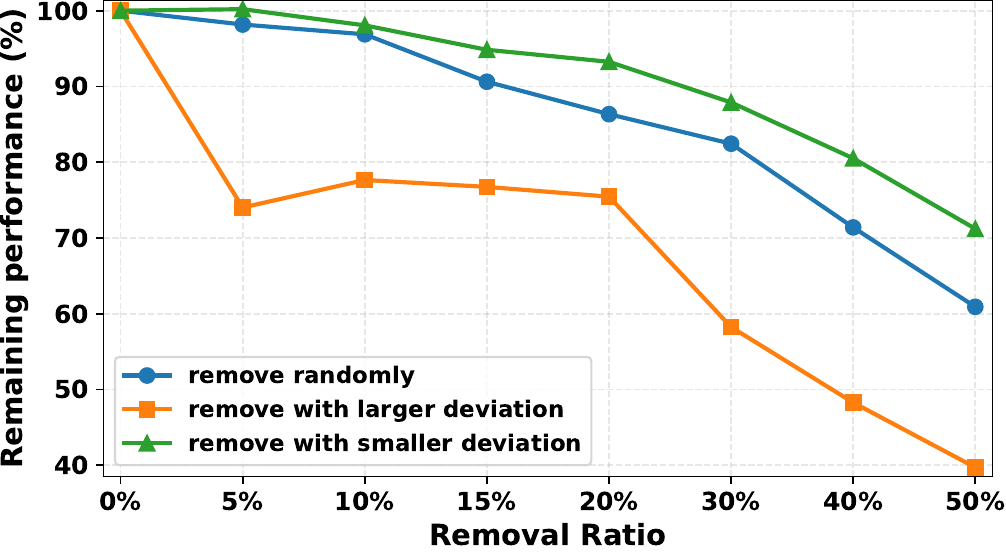}
    \vspace{-2mm}
    \caption{The impact on model performance of prioritizing the removal of KV pairs under different removal ratios. We find discarding KV pairs with larger deviations leads to faster performance drop than random removal or removing small-deviation pairs. Experiments are conducted on MileBench with Qwen2.5-VL-7B.}
    \label{fig22}
\vspace{-7mm}
\end{figure}

In this work, we revisit the problem from the perspective of the data distribution of the visual KV matrices themselves. Inspired by the common use of frequency-domain analysis in image processing~\cite{wallace1991jpeg}, we transform visual KV matrices into the frequency domain and analyze their energy distribution (i.e., the average power spectrum). As shown in Fig.~\ref{fig1}, we observe that the frequency-domain energy of KV matrices is predominantly concentrated at low frequencies, with high-frequency components occupying a relatively small proportion. Further, noting that removing outliers in model quantization often causes steep performance drops~\cite{dettmers2022gpt3,xiao2023smoothquant,frantar2022gptq}, we hypothesize that a similar outlier phenomenon may exist for KV Cache compression. To investigate, we apply a low-pass filter to obtain smoothed and dominant KV matrices (termed as \textit{Base KV}), then measure the deviation of the original KV matrices from \textit{Base KV} and remove KV pairs with different deviations.
We find that, discarding KV pairs with larger deviations leads to faster performance drop than random removal or removing small-deviation pairs (Fig.~\ref{fig22}), whereas discarding small-deviation KV pairs has smaller impact on performance. These results indicate KV pairs that differ substantially from the \textit{Base KV} are more likely to carry features critical to model inference, which we define as \textbf{Outlier KVs}.

Building on this novel finding, we propose \textbf{FlashCache}, a frequency-domain-guided Outlier-KV-aware compression approach with two core components. First, we introduce an Outlier KV Recognition Module that efficiently selects Outlier KVs in each layer. Concretely, we apply Discrete Cosine Transform (DCT)~\cite{ahmed2006discrete} to map the KV matrices to the frequency domain and retain only the dominant low-frequency components; we then perform the inverse DCT to return to the time domain and obtain the smoothed \textit{Base KV}; after that, we compute the mean squared error between each original KV pair and its \textit{Base KV} counterpart to select Outlier KVs. Second, we further observe differences in the low-frequency concentration phenomenon across each layer of the model in Fig.~\ref{fignn}, and propose a frequency-guided Dynamic Budget Allocation Module to dynamically determine the KV Cache size per layer. For each layer, we analyze the KV matrices in the frequency domain and compute the ratio of outlier-information energy to the total KV energy; we normalize these ratios into weights and allocate distinct per-layer retention quotas under a global cache budget. Across multiple models and datasets, FlashCache reduces KV usage and accelerates decoding under the same context length while preserving model performance. 

Our contributions are as follows:

\begin{itemize}
    \item To the best of our knowledge, we are the first to analyze multimodal KV Cache compression from a frequency-domain perspective. First, we find that the frequency energy of original KV matrices concentrates primarily in low frequency. Further, by applying a low-pass filter to obtain a smoothed \textit{Base KV}, we observe that preferentially discarding KV pairs that differ most from the \textit{Base KV} causes rapid performance degradation. It indicates that such KV pairs are more likely to encode features crucial for inference, which we define as \textbf{Outlier KVs}.
    \item Based on this novel finding, we propose \textbf{FlashCache}: a frequency-domain-guided Outlier-KV-aware framework for multimodal KV Cache compression. Specifically, Outlier KV Recognition Module is introduced to effectively identify Outlier KV. Then, Dynamic Budget Allocation Module is designed to dynamically allocate quotas across layers (based on the ratio of the frequency-domain energy intensity of outlier information to the total energy intensity.) to preserve more Outlier KVs. 
    \item As the first attention-score-free and training-free multimodal KV Cache compression framework, FlashCache is inherently compatible with efficient attention implementations (e.g., FlashAttention). We comprehensively validate FlashCache's effectiveness across multiple multimodal large language models and long-context benchmarks. Results demonstrate that FlashCache outperforms state-of-the-art multimodal KV Cache compression methods, delivering up to 1.69× faster decoding with 80\% lower KV memory while maintaining task performance.

\end{itemize}

\section{Related Works}
\label{sec:Related_Works}

\subsection{Multimodal Large Language Models}
\label{subsec:Multimodal Large Language Models}
Recently, multimodal large language models (MLLMs) have demonstrated powerful image-text understanding capabilities~\cite{liu2024mmbench,chen2024internvl,zhang2024long,li2024mini,ye2025dynamic,huang2024emr}. LLaVA~\cite{liu2024improved} aligns a visual encoder with a large language model via visual instruction tuning to enable general image-to-text understanding and multi-turn reasoning. InternVL~\cite{chen2024internvl} couples a strong vision backbone with a language model, introducing multi-granularity visual tokens and high-resolution modeling for improved detail perception and cross-task transfer. QwenVL~\cite{wang2024qwen2} extends the Qwen LLM with vision, supporting multi-image inputs, OCR, and referring expression grounding to strengthen open-domain multimodal reasoning. However, as context length increases (e.g., multi-image, high-resolution and video settings~\cite{song2024milebench,meng2024mmiu,hrbench,tu2025favor}), the multimodal KV Cache of MLLMs grows accordingly, leading to excessive GPU memory consumption and slower decoding speeds. Therefore, reducing the size of the KV Cache generated during long-context multimodal inference to accelerate decoding without compromising performance is both a necessary and valuable challenge.

\subsection{Visual Compression for MLLMs}
\label{subsec:Visual Compression for MLLMs}
In long-context scenarios, visual information experiences explosive growth. On one hand, in terms of volume, visual tokens often exceeds textual tokens by hundreds or even thousands of times~\cite{chen2024image}. On the other hand, visual signals are inherently sparser in information compared to texts produced by humans~\cite{marr2010vision}. Therefore, it is necessary to compress visual information during inference process of long-context MLLMs. Existing methods for compressing visual information in long-context MLLMs mainly include: visual token compression~\cite{chen2024image,tan2025tokencarve,zhang2024sparsevlm,shang2025llava,tan2025think}, quantization~\cite{yu2025mquant,xie2024advancing}, and multimodal KV Cache compression~\cite{li2024survey,he2025zipvl}. 

Among these, the KV Cache compression method has garnered significant attention for its ability to maintain performance while achieving high compression ratios, thereby reducing memory usage and accelerating decoding. LOOK-M~\cite{wan2024look} compresses the KV Cache by screening and merging visual KV pairs with low attention score once during the prefill stage. MEDA~\cite{wan2025meda} uses cross-modal attention entropy to allocate an appropriate KV Cache size per MLLM layer. However, these methods are not compatible with established efficient attention kernels ( i.e., FlashAttention~\cite{dao2022flashattention}) and overlook the contribution of value vectors, which directly influence the attention output. Therefore, we propose FlashCache, a frequency-domain-guided, training-free hierarchical KV compression framework for efficient multimodal long-context inference.

\label{sec:Method}

\begin{figure*}
    \centering
    \includegraphics[width=0.9\linewidth]{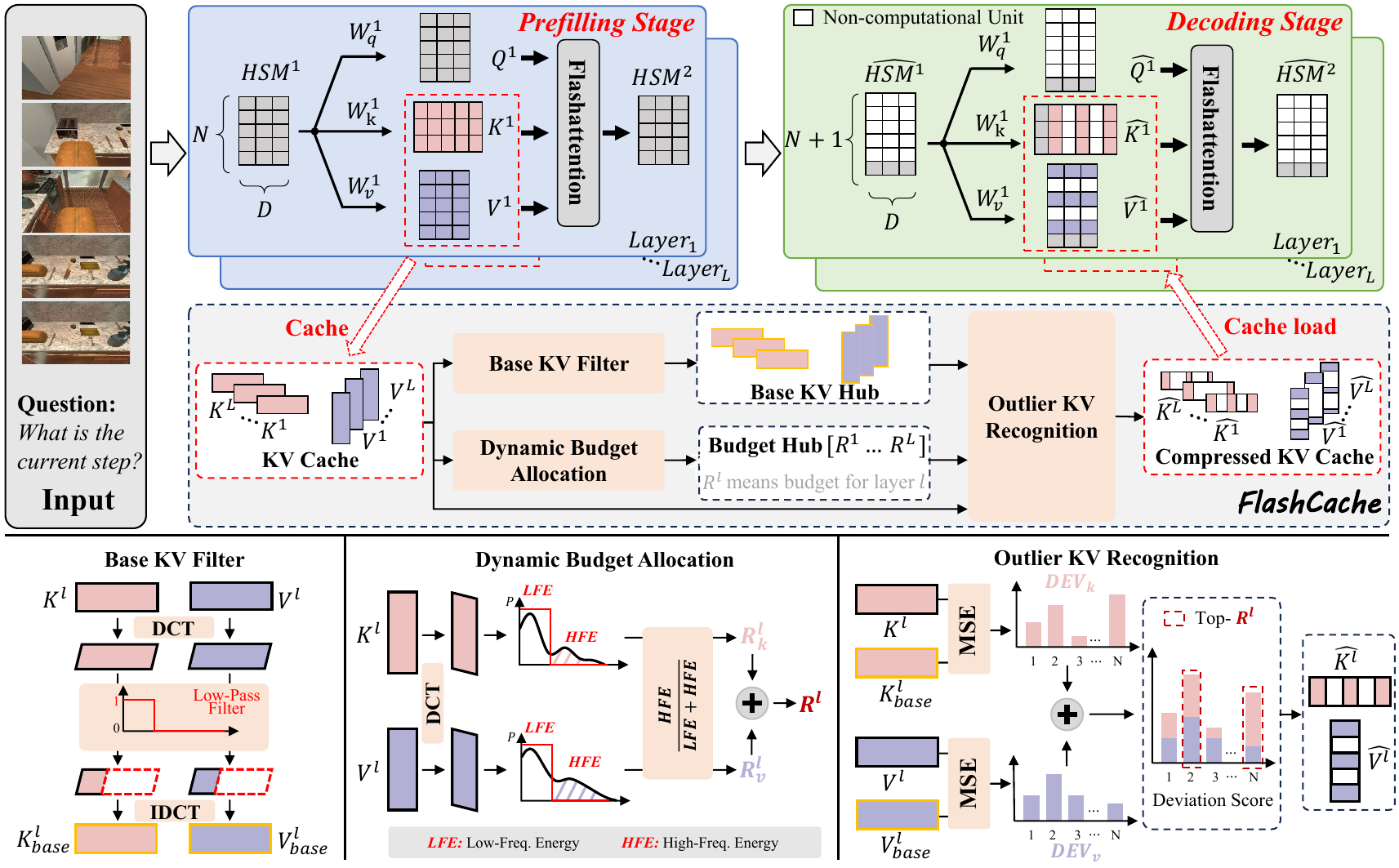}
    \caption{Overview of FlashCache. We perform compression on the multimodal KV Cache in a single operation after the prefilling stage. FlashCache comprises two core components: the Outlier KV Recognition Module and the Dynamic Budget Allocation Module. First, the Outlier KV Recognition Module obtains the primary smoothed component Base KV from the KV matrices via the Base KV Filter. Next, it 
     prioritizes retaining KV pairs with larger deviations of each KV pair from Base KV, which are defined as Outlier KVs.
    The Dynamic Budget Allocation Module measures the energy intensity of outlier information at each layer to dynamic allocation of the KV Cache size for each layer. ($HSM^{l}$ represents the hidden state matrix
 of layer $l$’s input in the model.)}
    \label{fig2}
\vspace{-1mm}
\end{figure*}

\section{Method}

\subsection{Preliminary}
\label{subsec:Preliminary}

In the existing MLLMs framework, KV Cache is a memory mechanism that is used to accelerate the inference process based on the attention model. The inference process of MLLMs based on KV Cache can be categorized into two stages: prefilling stage and decoding stage.

\textbf{Prefilling stage:} At this stage, MLLMs encodes the input sequence $X=[x_{0}^{T},x_{1}^{T},...,x_{N-2}^{V},x_{N-1}^{V}] \in \mathbb{R}^{N\times D}$, where $N$ represents the length of the input sequence, $D$ is the hidden dimension of the model. and $x^{T}$,$x^{V}$ represent text token and visual token respectively. To establish efficient contextual understanding, the Key and Value tensors are computed as:
\begin{align}
    K=XW_{k}, V=XW_{v}
\end{align}
where $W_{k},W_{v}\in \mathbb{R}^{D\times D}$ are the key and value projection matrices. Subsequently, the Key and Value tensors of the model's each layer is stored in the KV Cache.

\textbf{Decoding stage:} In the decoding stage, the KV cache is utilized and updated to generate tokens sequentially, avoiding recalculation of past KVs. At each time step t, the model computes only the Query, Key and Value tensors for the current token:
\begin{align}
    Q_{t}=x_{t}W_{q}, K_{t}=x_{t}W_{k}, V_{t}=x_{t}W_{v},
\end{align}
where $W_{q}\in \mathbb{R}^{D\times D}$ are the Query projection matrices.Subsequently, new KV pairs are added to the KV Cache:
\begin{align}
    K=[K,K_{t}],V=[V,V_{t}]
\end{align}

The output for the newly generated token is then computed as:
\begin{align}
    x_{out}=\text{Softmax}(Q_{t}K^{T}/\sqrt{D})V
\end{align}

In the long context multimodal models, the multimodal KV Cache grows as the input length increases in the long context setting, resulting in an increase in memory usage and a slowdown in decoding. Therefore, there is a strong need to reduce the multimodal KV Cache size to optimize the inference process.



\subsection{FlashCache}
\label{subsec:FlashCache}

As described in Sec.~\ref{sec:intro}, we find that KV pairs that differ significantly from the distribution of the main features of the original KV matrices are more likely to carry retrieval features that are important for model inference, which we define as Outlier KV. In this section, we propose a frequency-domain-guided, Outlier-KV-aware multimodal KV cache compression framework: FlashCache. FlashCache recognizes the Outlier KV of each layer and dynamically allocates a KV Cache budget for each layer of the model based on the energy strength of the outlier information. The overall framework is shown in Fig.~\ref{fig2}.

\subsubsection{Outlier KV Recognition Module}
\label{subsubsec:Outlier KV Recognition Module}

Following the prefilling stage, we obtain the KV Cache used by the model for decoding stage, as shown in Fig.~\ref{fig2}. For the given Key and Value tensors:
\begin{equation}
    \begin{aligned}
        K^{l}&=[k_{0}^{l},k_{1}^{l},...,k_{N-1}^{l}] \\
        V^{l}&=[v_{0}^{l},v_{1}^{l},...,v_{N-1}^{l}]
    \end{aligned}
\end{equation}
where $l$ represents model layer numbers, we find that their frequency domain energy intensity is concentrated at low frequency as shown in Fig.~\ref{fig1}. In natural signals, a large portion of the total energy is often concentrated in lower-frequency components, which correspond to the smooth, structural information. According to Parseval’s theorem~\cite{iwasaki2019deriving}, the energy of a signal is invariant between spatial and frequency domains. Therefore, we argue that retaining the principal components of the KV matrices in the frequency domain preserves the fundamental trends in KV representation. We therefore perform the low-pass filter of the original Key and Value tensors to obtain a smooth KV representation that is defined as \textit{Base KV}, which is able to characterize the primary trends in the KV matrices.

Specifically, we first transform original Key and Value tensors to the frequency domain using the Discrete Cosine Transform (DCT):
\begin{equation}
\begin{aligned}
    & C_{k}^{l}[m]=\alpha[m]\sum_{i=0}^{N-1}K^{l}[i]\cdot\phi_{N}(m,i) \\
    & C_{v}^{l}[m]=\alpha[m]\sum_{i=0}^{N-1}V^{l}[i]\cdot\phi_{N}(m,i) 
\end{aligned}
\end{equation}


where $m\in \{0,1,...,N-1\}$. $C$ is the frequency representation of original input. The basis function $\phi_{.}(.,.)$ and the normalization factor $\alpha[m]$ are defined as:
\begin{equation}
\begin{aligned}
  \phi_N(m,i) &= \cos\!\left(\frac{\pi}{N}\, m\left(i+\tfrac12\right)\right),\\
  \alpha[m] &= 
  \begin{cases}
    \sqrt{\frac{1}{N}}, & \text{if } m=0,\\[4pt]
    \sqrt{\frac{2}{N}}, & \text{otherwise}.
  \end{cases}
\end{aligned}
\end{equation}


As shown in Fig.~\ref{fig1}, we find that $K^{l},V^{l}$ frequency domain energy intensity is concentrated at low frequency. Based on this phenomenon, we perform the low-pass filtering approximation of $K^{l},V^{l}$:
\begin{equation}
\begin{aligned}
    \hat{C}^{l}[m]&=
    \begin{cases}
        C^{l}[m], & 0\leq m\leq \omega ,\\[4pt]
        0, & \omega\textless m\leq N-1
    \end{cases}
\end{aligned}
\end{equation}
where $ \omega= \gamma\cdot N $ and $\gamma$ represents the low-pass filter cutoff factor. Then we perform the Inverse Discrete Cosine Transform (IDCT) on the filtered frequency representation to obtain a smooth KV representation, named as \textit{Base KV}:
\begin{equation}
\begin{aligned}
    K_{base}^{l}[x]=\sum_{j=0}^{N-1}\alpha[j]\cdot \hat{C}_{k}^{l}[j]\cdot \phi_{N}(j,x) \\
    V_{base}^{l}[x]=\sum_{j=0}^{N-1}\alpha[j]\cdot \hat{C}_{v}^{l}[j]\cdot \phi_{N}(j,x)
\end{aligned}
\end{equation}
where $x\in[0,N-1]$ and is the position number of Key and Value tensors. We argue that constructing the \textit{Base KV} in the frequency domain by preserving dominant energy components is a principled information-retention strategy. Next we apply Mean Square Error (MSE) to calculate the deviation of original Key and Value tensors from \textit{Base KV}:
\begin{equation}
\begin{aligned}
     Dev_{k}[x]&= \text{MSE}(K^{l}[x],K_{base}^{l}[x]) \\
     Dev_{v}[x]&= \text{MSE}(V^{l}[x],V_{base}^{l}[x]) \\
     Dev[x]&=Dev_{k}[x]+Dev_{v}[x]
\end{aligned}
\end{equation}

Finally, we keep the KV pairs that deviate more in each layer according to the budget, which we denote as \textit{Outlier KV}.

\begin{figure}
    \centering
    \includegraphics[width=\linewidth]{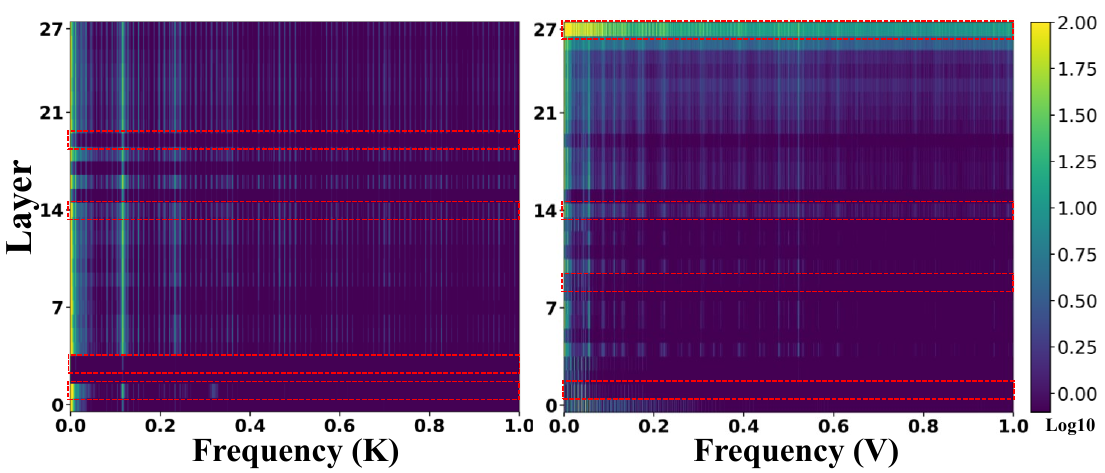}
    \caption{Frequency-domain energy distribution of KV matrices across different layers. We observe significant differences in the frequency-domain energy distribution of KV matrices across different layers. The red dashed box indicates the energy distributions of KV matrices of different layers show different modes. Experiments are conducted on MileBench~\cite{song2024milebench} with Qwen2.5-VL-7B.}
    \label{fignn}
    \vspace{-3mm}
\end{figure}

\begin{table*}[ht]
\centering
\caption{Performance of various KV cache strategies on several MLLMs on MileBench's tasks with KV Cache retention ratio $\rho=0.2$. Task T represents Temporal Multi-Image task. Task S represents Semantic Multi-Image task. NH represents Needle in a Haystack task. IR represents Image Retrieval task. The best results are highlighted in bold.}
\label{tab: milebench}
\small
\resizebox{\linewidth}{!}{
\renewcommand{\arraystretch}{1.0} 
\begin{tabular}{l|cccc|cccc|cccc}
\toprule
\multirow{2}{*}{Method} & \multicolumn{4}{c|}{LLaVA-OneVision-1.5-8B} & \multicolumn{4}{c|}{Qwen2.5-VL-7B} & \multicolumn{4}{c}{Qwen2.5-VL-32B} \\
\cmidrule{2-13}
& Task T & Task S & NH & IR & Task T & Task S & NH & IR & Task T & Task S & NH & IR \\
\midrule
\rowcolor{gray!16}
\textbf{Full Cache} & 46.97 & 65.33 & 27.03 & 11.67 & 55.59 & 69.17 & 27.35 & 14.17 & 56.51 & 65.21 & 27.19 & 18.17 \\
\midrule
\textbf{StreamingLLM~\cite{xiao2023efficient}} & 46.97 & 65.04 & 9.22 & 12.17 & 55.59 & 67.51 & 9.69 & 14.00 & 56.75 & 63.15 & 3.75 & 19.33 \\
\textbf{H2O~\cite{zhang2023h2o}} & 46.97 & 65.29 & 19.69 & 12.33 & 55.59 & 68.60 & 12.66 & 14.67 & 57.01 & 64.73 & 14.32 & 18.00 \\
\textbf{SnapKV~\cite{li2024snapkv}} & 46.97 & 65.27 & 19.37 & 12.17 & 55.59 & 68.27 & 13.59 & 15.33 & 57.07 & 65.61 & 22.34 & 18.83 \\
\textbf{LOOK-M~\cite{wan2024look}} & 46.97 & 65.23 & 18.60 & 11.83 & 55.55 & 67.50 & 11.88 & 11.83 & 51.03 & 53.37 & 5.78 & 13.00 \\
\textbf{MEDA~\cite{wan2025meda}} & 46.91 & 65.08 & 19.97 & 12.33 & 55.59 & 68.13 & 9.07 & 14.50 & 51.86 & 54.01 & 5.16 & 14.50 \\
\rowcolor{Lavender!16}
\textbf{FlashCache}  & \textbf{46.97} & \textbf{65.46} & \textbf{25.94} & \textbf{12.83} & \textbf{55.59} & \textbf{68.85} & \textbf{26.72} & \textbf{15.50} & \textbf{57.30} & \textbf{66.21} & \textbf{24.69} & \textbf{19.50} \\
\bottomrule
\end{tabular}
}
\vspace{-1mm}
\end{table*}
\begin{figure*}[t]
    \centering
    \includegraphics[width=\linewidth]{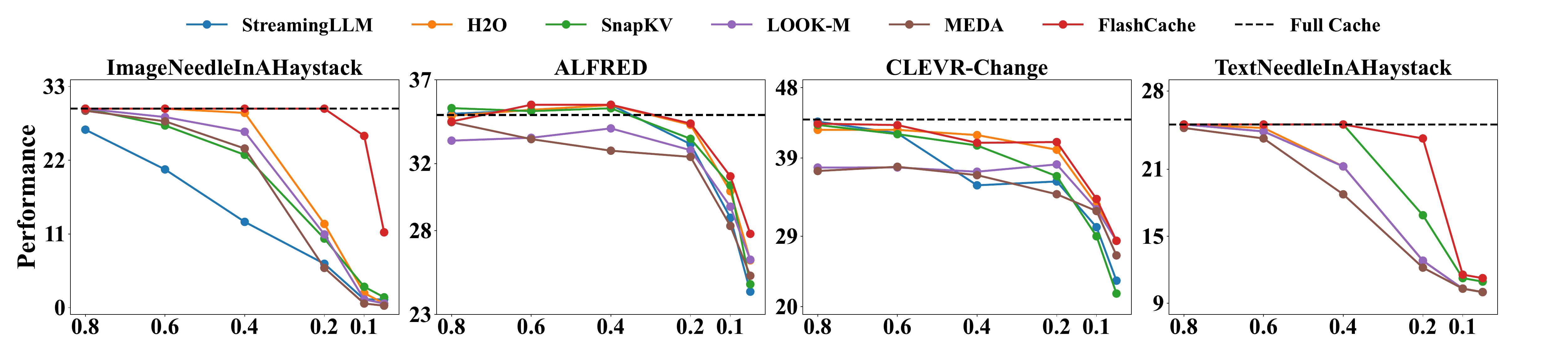}
    \caption{The impact of KV Cache retention ratio changes on performance. We compare the performance impact of different KV Cache compression methods across six KV Cache retention ratio settings (0.8, 0.6, 0.4, 0.2, 0.1, 0.05), where FlashCache consistently shows superior performance, particularly under low retention ratios. The horizontal axis represents the KV Cache retention ratio.}
    \label{fig:ab_ratio}
\vspace{-1mm}
\end{figure*}

\subsubsection{Dynamic Budget Allocation Module}
\label{subsubsec:Dynamic Budget Allocation Module}
Previous studies~\cite{zhang2024sparsevlm,cai2024pyramidkv,li2024survey} have found that different layers of MLLMs exhibit varying degrees of information redundancy. Allocating distinct KV Cache budgets to each layer maximizes the reduction of information loss caused by compression. 
At the same time, we observe that the low-frequency concentration in the frequency domain of the KV matrices across different layers of the model is not consistent, as shown in Fig.~\ref{fignn}. In particular, different transformer layers exhibit varying proportions of outlier energy: in the frequency domain, the key-value matrices of certain layers are predominantly concentrated in low-frequency principal components, while other layers carry relatively more high-frequency outlier energy. This cross-layer disparity suggests that a uniform KV retention ratio for all layers would be suboptimal. Based on this, we further propose a frequency-domain guided Dynamic Budget Allocation Module to dynamically determine the size of the KV Cache for each layer.

First, we calculate the frequency-domain energy distribution of the original Key and Value tensors. The use of an orthonormal DCT satisfies Parseval’s theorem, enabling us to directly compute the frequency-domain power spectrum from the transform coefficients:
\begin{equation}
\begin{aligned}
    P_{k}^{l}[m]&=|C_{k}^{l}[m]|^{2} \\
    P_{v}^{l}[m]&=|C_{v}^{l}[m]|^{2}
\end{aligned}
\end{equation}

We compute the proportion of outlier information energy intensity in original Key and Value tensors for each layer relative to their total energy intensity:
\begin{equation}
\begin{aligned}
    R_{k}^{l}&=(\sum_{\ell=\omega+1}^{N-1}P_{k}^{l}[\ell])/(\sum_{\ell=0}^{N-1}P_{k}^{l}[\ell]) \\
    R_{v}^{l}&=(\sum_{\ell=\omega+1}^{N-1}P_{v}^{l}[\ell])/(\sum_{\ell=0}^{N-1}P_{v}^{l}[\ell]) \\
    R^{l}&=R_{k}^{l}+R_{v}^{l}
\end{aligned}
\end{equation}

Finally, we normalize each layer's ratio $R=[R^{1},R^{2},...,R^{l}]$ into weights and allocate different retention quotas to each layer under the constraint of a given overall cache ratio. Finally, we allocate the corresponding share of the KV budget to each layer based on this ratio for decoding.

\section{Experiments}
\label{sec:Experiments}

\subsection{Experimental Setup}
\label{subsec:Experimental Setup}
\textbf{Datasets and models.} 
To validate the effectiveness of our approach, we evaluate FlashCache on diverse benchmarks spanning multiple tasks:
three multi-image understanding benchmarks (MileBench~\cite{song2024milebench},  MUIRBench~\cite{wang2024muirbench} and MMMU~\cite{yue2023mmmu}), two high-resolution benchmarks (V*~\cite{vstar} and HR-Bench~\cite{hrbench}) and one video benchmark (FAVOR-Bench~\cite{tu2025favor}). 
We verify FlashCache on three
MLLMs: LLaVA-OneVision-1.5-8B-Instruct~\cite{LLaVAOneVision15} and Qwen2.5-VL-7B/32B-Instruct~\cite{wang2024qwen2}.

\textbf{Implementation details.} 
All experiments are conducted on a single NVIDIA H200 (141GB) GPU with FlashAttention~\cite{dao2022flashattention}. 
We compare FlashCache against state-of-the-art KV cache eviction methods:
StreamingLLM~\cite{xiao2023efficient}, H2O~\cite{zhang2023h2o}, SnapKV~\cite{li2024snapkv}, LOOK-M~\cite{wan2024look}, and MEDA~\cite{wan2025meda}. 
We utilize NVIDIA's CuPy operator library to accelerate the DCT operation within FlashCache.

\subsection{Experiments on Multi-Images Understanding Benchmarks}


\textbf{Experiments on Milebench.} 
We first evaluate FlashCache's dynamic compression capabilities in multi-image scenarios using MileBench across models of varying architectures and scales. 
We set the  KV Cache retention ratio to $\rho=0.2$  to compare the performance of different algorithms under low compression settings. 
As shown in Tab.~\ref{tab: milebench}, FlashCache outperforms other methods on most tasks without requiring attention score recalculation. This demonstrates that we can identify and preserve important Key-Value pairs solely based on the distribution characteristics of the Key-Value matrices. 
Particularly in the NH task, FlashCache demonstrates a significant advantage, indicating its effectiveness in preserving essential Key-Value pairs for subsequent retrieval in multimodal long-term contexts. 

Fig.~\ref{fig:ab_ratio} details model performance across different KV cache retention ratios on four datasets included in MileBench, 
covering a diverse set of vision-language tasks. 
We employ Qwen2.5-VL-7B as the backbone model. For each dataset, we vary the KV Cache retention ratio across a wide range of settings, from high-retention to highly compressed configurations, and measure the resulting degradation in task performance. 
The experimental results show that, as the KV cache retention ratio decreases, FlashCache consistently outperforms all competing methods, exhibiting slower performance degradation and better robustness in low-retention regimes. These observations indicate FlashCache's superior ability to preserve informative KV pairs, thereby enabling efficient memory usage while maintaining strong downstream performance. 
Additional experiments with lower KV Cache retention ratios are provided in the Appendix, which demonstrate more pronounced advantages of our method.

\newcolumntype{C}{>{\centering\arraybackslash}X} 

\begin{table}[t]
\centering
\caption{{Performance of various KV Cache strategies on Qwen2.5-VL-7B-Instruct on MUIRBench and MMMU with different KV Cache retention ratios. The best results are highlighted in bold.}}
\label{tab: muir_mmmu}
\small
\scalebox{0.98}{
\begin{tabularx}{0.48\textwidth}{{l|CC|CC}}
\toprule
Method & \multicolumn{2}{c}{MUIRBench}&  \multicolumn{2}{c}{MMMU}  \\ 
\midrule
\rowcolor{gray!16}
\textbf{Full Cache} & \multicolumn{2}{c}{45.54} & \multicolumn{2}{c}{53.18}  \\
\midrule
\textbf{retention ratio $\rho$} & $0.1$ & $0.05$ & $0.1$ & $0.05$  \\
\midrule
\textbf{StreamingLLM~\cite{xiao2023efficient}}           & 37.58 & 37.58 & 50.0 & 50.0  \\
\textbf{H2O~\cite{zhang2023h2o}}           & OOM & OOM  & 53.18 & 51.18 \\
\textbf{SnapKV~\cite{li2024snapkv}}           & OOM & 44.42 & 53.18 & 52.18 \\
\textbf{LOOK-M~\cite{wan2024look}}           & OOM & OOM & 53.18 & 52.18 \\
\textbf{MEDA~\cite{wan2025meda}}           & OOM & OOM  & 53.18 & 51.36 \\
\rowcolor{Lavender!16}
\textbf{FlashCache} & \textbf{44.42} & \textbf{44.42} & \textbf{53.18} & \textbf{52.27} \\
\bottomrule
\end{tabularx}
}
\vspace{-1mm}
\end{table}
\newcolumntype{C}{>{\centering\arraybackslash}X} 

\begin{table}[t]
\centering
\caption{{Performance of various KV Cache strategies on Qwen2.5-VL-7B-Instruct on high resolution benchmarks (V* and HR-Bench) with different KV Cache retention ratios. The best results are highlighted in bold.}}
\label{tab: high_resolution}
\small
\scalebox{0.98}{
\begin{tabularx}{0.48\textwidth}{{l|CC|CC}}
\toprule
Method & \multicolumn{2}{c}{V*}&  \multicolumn{2}{c}{HR-Bench}  \\ 
\midrule
\rowcolor{gray!16}
\textbf{Full Cache} & \multicolumn{2}{c}{80.23} & \multicolumn{2}{c}{70.75}  \\
\midrule
\textbf{retention ratio $\rho$} & $0.1$ & $0.05$ & $0.1$ & $0.05$  \\
\midrule
\textbf{StreamingLLM~\cite{xiao2023efficient}}           & 78.77 & 78.65 & 71.25 & 71.25  \\
\textbf{H2O~\cite{zhang2023h2o}}           & 78.77 & 78.77  & 70.75 & 71.12 \\
\textbf{SnapKV~\cite{li2024snapkv}}           & 79.56 & 78.89 & 70.88 & 71.12 \\
\textbf{LOOK-M~\cite{wan2024look}}           & 78.31 & 77.78 & 70.0 & 70.25 \\
\textbf{MEDA~\cite{wan2025meda}}           & 78.84 & 77.78  & OOM & OOM \\
\rowcolor{Lavender!16}
\textbf{FlashCache} & \textbf{80.23} & \textbf{79.66} & \textbf{71.25} & \textbf{72.38} \\
\bottomrule
\end{tabularx}
}
\vspace{-3mm}
\end{table}


\textbf{Experiments on MUIRBench and MMMU.}
We further validate FlashCache's versatility on MUIRBench and MMMU benchmarks using Qwen2.5-VL-7B with retention ratios of 0.1 and 0.05. 
Tab.~\ref{tab: muir_mmmu} reveals that full-attention computation methods encounter out-of-memory issues on MUIRBench when processing longer input sequences, highlighting FlashCache's efficiency and the importance of compatibility with FlashAttention.
Our approach achieves state-of-the-art performance on both benchmarks under two KV retention ratio settings, demonstrating its effectiveness in general multi-image scenarios.

\subsection{Experiments on High-Resolution Benchmarks}
\label{subsec:Experiments on Experiments on High-Resolution Benchmarks}

To further evaluate FlashCache's effectiveness in long-context scenarios, we conduct experiments on two high-resolution benchmarks (V* and HR-Bench), as shown in Tab.~\ref{tab: high_resolution}. In the HR-Bench evaluation, we use the Qwen2-32B model to extract and assess whether the outputs of MLLMs are correct. 
Specifically, we employ Qwen2.5-VL-7B for the experiment with retention ratios of 0.1 and 0.05 and present the accuracy metrics for different compression methods. 
Under low KV cache retention ratios, FlashCache achieves minimal performance degradation across both benchmarks. In certain experimental configurations, FlashCache even matches or surpasses full cache performance, demonstrating the robustness of our approach and its generalizability in long-context scenarios.
\begin{table}[t]
\centering
\caption{\small{Performance of various KV Cache strategies on Qwen2.5-VL-7B-Instruct on FAVOR-Bench with different KV Cache retention ratios. The best results are highlighted in bold.}}
\label{tab: favor}
\footnotesize
\scalebox{0.95}{
\begin{tabularx}{0.5\textwidth}{l|*{8}{>{\centering\arraybackslash}X}}
\toprule
Method & AS & HAC & SAD & MAD & CM & NSM & \textbf{\textit{all}} \\ 
\midrule
\rowcolor{gray!16}
\textbf{Full Cache} & 39.63 & 43.61 & 43.2 & 43.57 & 33.67 & 40.62 & \textit{40.91}  \\
\midrule
\multicolumn{8}{c}{KV Cache retention ratio $\rho=0.1$} \\
\midrule
\textbf{StreamingLLM~\cite{xiao2023efficient}}           & 30.94 & 39.13 & 39.05 & 40.5 & 31.35 & 28.12 & \textit{35.57}  \\
\textbf{H2O~\cite{zhang2023h2o}}           & \textbf{32.16} & 38.22 & 39.05 & 40.66 & 31.35 & 23.44 & \textit{35.78}  \\
\textbf{SnapKV~\cite{li2024snapkv}}           & 28.9 & 39.71 & 40.91 & 39.83 & 33.49 & \textbf{39.06} & \textit{35.67}  \\
\textbf{LOOK-M~\cite{wan2024look}}           & 30.38 & \textbf{40.23} & 39.53 & 43.4 & 32.56 & 25.0 & \textit{36.25}  \\
\textbf{MEDA~\cite{wan2025meda}}           & 26.17 & 36.28 & 37.61 & 39.75 & 31.91 & 21.88 & \textit{33.11}  \\
\rowcolor{Lavender!16}
\textbf{FlashCache}  & 30.38 & 40.1 & \textbf{41.16} & \textbf{40.91} & \textbf{34.14} & 35.94 & \textit{\textbf{36.49}} \\
\midrule
\multicolumn{8}{c}{KV Cache retention ratio $\rho=0.05$} \\
\midrule
\textbf{StreamingLLM~\cite{xiao2023efficient}}           & 20.52 & 29.4 & 32.55 & 31.29 & 28.0 & 12.5 & \textit{27.14}  \\
\textbf{H2O~\cite{zhang2023h2o}}           & 21.35 & 28.68 & 32.19 & 31.45 & 28.56 & 12.5 & \textit{27.3}  \\
\textbf{SnapKV~\cite{li2024snapkv}}           & \textbf{24.19} & 32.25 & 34.66 & 34.69 & 29.02 & 15.62 & \textit{29.95}  \\
\textbf{LOOK-M~\cite{wan2024look}}           & 20.29 & 30.11 & 32.49 & 33.61 & 31.81 & 12.5 & \textit{28.03 } \\
\textbf{MEDA~\cite{wan2025meda}}           & 15.17 & 24.72 & 25.99 & 29.13 & 27.53 & 12.5 & \textit{22.83}  \\
\rowcolor{Lavender!16}
\textbf{FlashCache}  & 21.27 & \textbf{34.78} & \textbf{37.55} & \textbf{34.69} & \textbf{33.21} & \textbf{26.56} & \textit{\textbf{30.71}} \\
\bottomrule
\end{tabularx}
}
\vspace{-2mm}
\end{table}
\begin{table*}[t]
\caption{Ablation Study on the low-pass filter cut-off factor $\gamma$. The range of values for $\gamma$ is from 0 to 1.}
\centering
\label{tab:ab_gamma}
\begin{tabular}{l|lllllllll}
\toprule
Cut-off Factor \textbf{$\gamma$}               & 0.1   & 0.2   & 0.3   & 0.4   & 0.5   & 0.6   & 0.7   & 0.8 & 0.9 \\ 
\midrule
ImageNeedleInAHaystack & 29.06 & 29.69 & 25.0  & 23.12 & 22.5  & 23.44 & 22.81 & 21.82   & 20.08   \\
GPR1200                & 15.0  & 15.5  & 15.17 & 15.0  & 15.17 & 14.33 & 14.83 & 14.32   & 13.05  \\
\bottomrule
\end{tabular}
\vspace{-1mm}
\end{table*}

\subsection{Experiments on Video Benchmark}
\label{subsec:Experiments on Video Benchmark}
We further evaluate FlashCache in video understanding scenarios by validating it on FAVOR-Bench with Qwen2.5-VL-7B, and report the corresponding results in Tab.~\ref{tab: favor}. The experimental results show that FlashCache consistently outperforms competitive baselines on a range of video comprehension tasks. In particular, even under very low KV cache retention ratios (e.g., 0.05), FlashCache retains a larger fraction of the original model’s performance while simultaneously delivering noticeable acceleration.

\subsection{Efficiency Analysis}
\label{subsec:Efficiency Analysis}
In this subsection, we evaluate the computational efficiency of FlashCache, which is a critical practical consideration.
All measurements are done using FlashAttention.

\begin{figure}
    \centering
    \includegraphics[width=\linewidth]{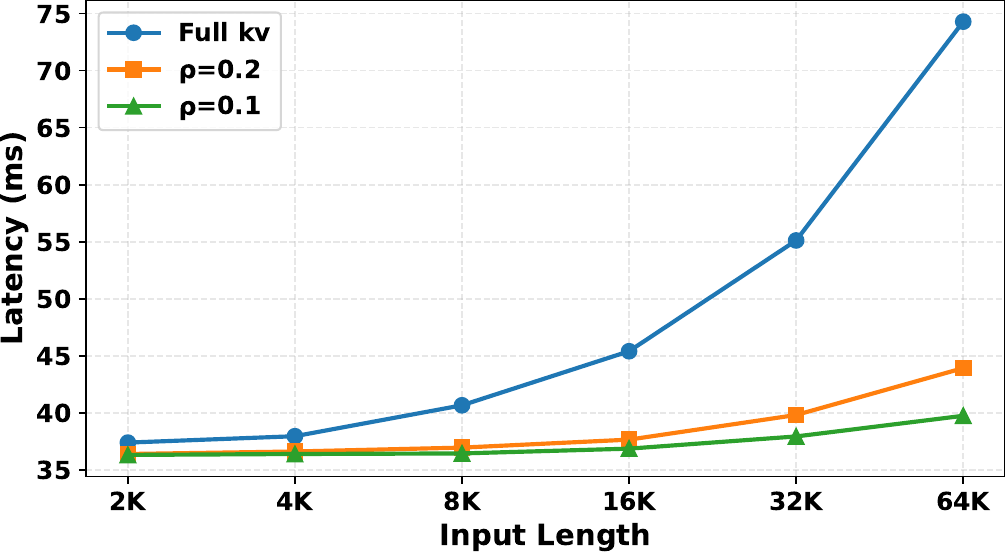}
    \caption{Decoding latency evaluation for FlashCache. Benefiting from the reduction in KV cache, FlashCache can maintain nearly constant decoding latency as input length increases. }
    \label{fig:dl}
\vspace{-3mm}
\end{figure}

\textbf{Decoding Latency.}
We measure the decoding latency during the generation of 100-token output sequences to obtain the per-token latency across input token lengths \{2K, 4K, 8K, 16K, 32K, 64K\}.
At KV retention ratios of 0.2 and 0.1, Fig.~\ref{fig:dl} demonstrates that the reduction in KV cache within FlashCache significantly lowers computational load during inference, thereby accelerating inference speed. 
Acceleration effects become more pronounced with longer input sequences, achieving up to a 1.69× speedup at the retention ratio of 0.2.

\textbf{Method Latency.}
We measure and compare additional overhead introduced by attention-based methods and FlashCache across input lengths \{2K, 4K, 8K\} in Tab.~\ref{tab:Method Latency}. 
Since SnapKV uses partial attention computation, its computational overhead is significantly lower than other methods based on attention computation.
LOOK-M and MEDA have significantly higher computational overhead due to the introduction of the KV merging technique. Instead, FlashCache prioritizes important KV pairs based on the distribution of KV matrices themselves, thus natively adapting to efficient attention mechanisms such as FlashAttention. 
We utilize the CuPy operator library developed by NVIDIA to further accelerate the implementation of the DCT within FlashCache.

\subsection{Ablation Study}
\label{subsec:Ablation Study}

\textbf{Ablation Study on the low-pass filter cut-off factor $\gamma$.}
In this section, we first examine the impact of $\gamma$ on FlashCache, which is used as a control parameter for the cutoff frequency of the low-pass filter in Sec.~\ref{sec:Method}.
Specifically, we employ Qwen2.5-VL-7B for the experiment, with the KV retention ratio set to 0.2 and the range of values for $\gamma$ from 0 to 1. 
To facilitate comparison of differences, we conduct experiments on the ImageNeedleInAHaystack and GPR1200 datasets (both included in MileBench). 
As shown in Tab.~\ref{tab:ab_gamma}, we observe that as $\gamma$ increases, the model's performance gradually declines. Optimal model performance is achieved when $\gamma$ falls within the range of 0.1 to 0.2. 
Since the KV matrix is predominantly concentrated in low-frequency ranges, excessive $\gamma$ values prevent the low-pass filter from effectively extracting the primary representation of the KV matrix (\textit{Base KV}). 
This failure to identify the Outlier KV consequently leads to a decline in model performance. This phenomenon is consistent with our core findings.

\textbf{The Effect of Dynamic Budget Allocation Module.}
~Dynamic Budget Allocation (DBA) module adaptively assigns KV retention budgets across layers according to the ratio of the frequency-domain energy of outlier information to the total energy, thereby preserving more \textit{Outlier KV}. To assess its contribution, we conduct an ablation study with Qwen2.5-VL-7B under a KV retention ratio of 0.2, and report the results in Tab.~\ref{tab:ab_dba}. Removing the DBA module leads to a clear degradation in performance, indicating that dynamically distributing the KV budget across layers is crucial for multimodal long-context scenarios.

%



\begin{table}[t]
\caption{Method Latency. We measure the additional time overhead introduced by attention-based selection methods and FlashCache for different input lengths. The metric given is in milliseconds (ms).}
\centering
\footnotesize
\label{tab:Method Latency}
\begin{tabular}{c|ccccc}
\toprule
Method & \makecell{H2O \\ \cite{zhang2023h2o}}     & \makecell{SnapKV \\ \cite{li2024snapkv}} & \makecell{LOOK-M \\ \cite{wan2024look}} & \makecell{MEDA \\ \cite{wan2025meda}}  & FlashCache \\
\midrule
2K     & 3.83   & 2.53   & 6.93   & 16.6  & \textbf{1.66}       \\
4K     & 10.29 & 4.95   & 18.66  & 38.39 & \textbf{3.86}       \\
8K     & 27.62  & 9.57   & 53.97  & 83.75 & \textbf{6.77}       \\
\bottomrule
\end{tabular}
\vspace{-3mm}
\end{table}


\begin{table}[t]
\caption{The effect of dynamic budget allocation module (DBA). 'INIAH' means ImageNeedleInAHaystack dataset.}
\centering
\small
\label{tab:ab_dba}
\resizebox{0.48\textwidth}{!}{
\begin{tabular}{c|cccc}
\toprule
Dataset & INIAH & GPR1200 & ALFRED & CLEVR-Change \\
\midrule
w/o DBA & 24.69 & 14.67 & 34.32 & 35.85\\
w DBA   & \textbf{29.69} & \textbf{15.50} & \textbf{34.39} & \textbf{41.04}\\
\bottomrule
\end{tabular}
}
\vspace{-4mm}
\end{table}

\section{Conclusion}
\label{sec:Conclusion}
In this work, we present FlashCache, a frequency-domain–guided, Outlier-KV-aware KV Cache compression framework that is inherently compatible with efficient attention implementations. 
First, we observe that KV pairs exhibiting significant deviations from the primary distribution of the KV matrices play a more critical role in model inference. We define these as Outlier KVs. FlashCache efficiently identifies and prioritizes the retention of these Outlier KVs. Furthermore, FlashCache dynamically allocates KV Cache sizes for each model layer by calculating the strength of outlier information within the KV matrices at each layer, thereby preserving more Outlier KVs. FlashCache is the first attention-score-free and training-free multimodal KV Cache compression framework, inherently compatible with efficient attention implementations (e.g., FlashAttention). Across multiple models and datasets, FlashCache outperforms state-of-the-art multimoal KV compression methods.

\section{Acknowledgments}
\label{sec:Acknowledgments}
This work is supported by National Key Research and Development Program of China (No. 2022ZD0160101), Shanghai Natural Science Foundation (No. 23ZR1402900), Shanghai Science and Technology Commission Explorer Program Project (24TS1401300), Shanghai Municipal Science and Technology Major Project (No.2021SHZDZX0103).
The computations in this research were performed using the CFFF platform of Fudan University.

{
    \small
    \bibliographystyle{ieeenat_fullname}
    \bibliography{main}
}
\clearpage
\setcounter{page}{1}
\appendix
\maketitlesupplementary

\section*{Outline for Supplementary Material}

\begin{outline}
\0 Due to the page limitation for the submission paper, we present additional details and visualization results from the following aspects:

\1 A: More Experiments in different KV Retention Ratios
\1 B: More Experiments on Decoding Stage KV Retention
\1 C: More Ablation Experiments on Low-Pass Cutoff Factor $\gamma$
\1 D: More Observations
\2 D.1: Frequency-domain energy distribution of the KV matrices
\2 D.2: Frequency-domain energy distribution of KV matrices across different layers
\2 D.3: The impact on model performance of prioritizing the removal of KV pairs under different removal ratios
\1 E: More Efficiency Analysis
\1 F: Case Study 
\1 G: Benchmarks
\2 G.1 Multi-Images Understanding Benchmarks 
\2 G.2 High Resolution Benchmarks
\2 G.3 Video Understanding Benchmark
\1 H: Models
\2 H.1 Qwen2.5-VL-7B-Instruct
\2 H.2 LLaVA-OneVision-1.5-8B-Instruct
\1 I: Limitation and Future Work
\end{outline}

\section{More Experiments in different KV Retention Ratios}
\label{sec:More Experiment in different KV Retention Ratios}
In this section, we present the performance of various models on MileBench under lower KV Cache retention rates in Tab.~\ref{supp_tab: milebench}. Task T represents Temporal Multi-Image task. Task S represents Semantic Multi-Image task. NH represents Needle in a Haystack task. IR represents Image Retrieval task. As shown in Tab.~\ref{supp_tab: milebench}, FlashCache outperforms other methods on most tasks without requiring attention score recalculation. This demonstrates that we can identify and preserve important Key-Value pairs solely based on the distribution characteristics of the Key-Value matrices. 

\section{More Experiments on Decoding Stage KV Retention}
Following established baselines (LOOK-M, MEDA), we focus on compressing prefill KV Cache while fully retaining decoding KVs. To validate robustness in long-generation scenarios, we evaluate \textit{FlashCache} on the MathVista(COT) benchmark. As shown in Tab.\ref{supp_tab:mathvista}, \textit{FlashCache} also maintains performance advantages in reasoning-heavy task.

\section{More Ablation Experiments on Low-Pass Cutoff Factor $\gamma$}
Additional ablation on LLaVA-OneVision-8B (Tab.\ref{supp_tab:ab_gamma}) aligns with our Qwen results, identifying $\gamma \in [0.1, 0.2]$ as optimal and robust. Theoretically, a lower $\gamma$ creates smoother \textit{Base KV} that forces the deviation term to capture critical high-frequency features. Conversely, a high $\gamma$ causes \textit{Base KV} to overfit, making outliers indistinguishable. Thus, $\gamma \in [0.1, 0.2]$ is a robust guideline across models and datasets.

\begin{table*}[ht]
\centering
\caption{Performance of various KV cache strategies on several MLLMs on MileBench's tasks with KV Cache retention ratio $\rho=0.1, 0.05$. The best results are highlighted in bold.}
\label{supp_tab: milebench}
\small
\resizebox{\linewidth}{!}{
\renewcommand{\arraystretch}{1.0} 
\begin{tabular}{l|cccc|cccc|cccc}
\toprule
\multirow{2}{*}{Method} & \multicolumn{4}{c|}{LLaVA-OneVision-1.5-8B} & \multicolumn{4}{c|}{Qwen2.5-VL-7B} & \multicolumn{4}{c}{Qwen2.5-VL-32B} \\
\cmidrule{2-13}
& Task T & Task S & NH & IR & Task T & Task S & NH & IR & Task T & Task S & NH & IR \\
\midrule
\rowcolor{gray!16}
\textbf{Full Cache} & 46.97 & 65.33 & 27.03 & 11.67 & 55.59 & 69.17 & 27.35 & 14.17 & 56.51 & 65.21 & 27.19 & 18.17 \\
\midrule
\rowcolor{gray!16}
\multicolumn{13}{c}{$\rho=0.1$} \\
\textbf{StreamingLLM~\cite{xiao2023efficient}} & 46.81 & 58.40 & 6.72 & 12.5 & 54.71 & 59.23 & 6.56 & \textbf{15.83} & 57 & 56.15 & 2.5 & 20.00 \\
\textbf{H2O~\cite{zhang2023h2o}} & 46.84 & 58.51 & 15.94 & 11.83 & 54.58 & 58.65 & 6.25 & 14.83 & 56.94 & 58.33 & 2.5 & 21.0 \\
\textbf{SnapKV~\cite{li2024snapkv}} & 46.97 & 57.62 & 10.47 & 12.67 & 55.58 & \textbf{61.30} & 7.19 & 15.50 & 57.06 & 58.71 & 16.06 & 18.83 \\
\textbf{LOOK-M~\cite{wan2024look}} & 46.84 & 58.51 & 15.94 & 11.83 & 55.08 & 58.54 & 5.78 & 13.67 & 51.91 & 58.99 & 2.81 & 24.67 \\
\textbf{MEDA~\cite{wan2025meda}} & 44.23 & 52.76 & 14.06 &12.17 & 54.24 & 57.05 & 5.47 & 13.67 & 51.31 & 48.45 &2.97 & \textbf{28.67} \\
\rowcolor{Lavender!16}
\textbf{FlashCache}  & \textbf{47.09} & \textbf{58.56} & \textbf{20.00} &\textbf{12.83} & \textbf{55.65} & 60.93 & \textbf{18.45} &14.5 & \textbf{57.23} & \textbf{59.08} &\textbf{16.10} & 20.17 \\
\midrule
\rowcolor{gray!16}
\multicolumn{13}{c}{$\rho=0.05$} \\
\textbf{StreamingLLM~\cite{xiao2023efficient}} & 46.52 & 57.05 & 5.16 & 12.00 & 49.46 & 53.64 & 5.63 & 15.00 & 51.94 & 42.34 & 1.25 & 26.83 \\
\textbf{H2O~\cite{zhang2023h2o}} & 47.01 & 56.99 & 12.50 & 11.50 & 50.47 & 52.77 & 5.31 & 11.33 & 52.31 & 48.44 & 1.25 & 23.67 \\
\textbf{SnapKV~\cite{li2024snapkv}} & 45.97 & \textbf{57.82} & 7.97 & 13.00 & 51.52 & \textbf{56.33} & 6.25 & 15.33 & 55.69 & 49.07 & 6.25 & 19.83 \\
\textbf{LOOK-M~\cite{wan2024look}} & \textbf{48.01} & 56.99 & 12.50 & 11.50 & 51.69 & 53.67 & 5.31 & \textbf{16.33} & 49.97 & 45.77 & 2.5 &\textbf{27.83} \\
\textbf{MEDA~\cite{wan2025meda}} & 38.40 & 48.25 & 10.31 & 13.00 & 51.74 & 52.95 & 5.16 & 15.33 & 51.86 & 50.01 & 2.65 & 26.33 \\
\rowcolor{Lavender!16}
\textbf{FlashCache}  & 45.51 & 51.12 & \textbf{15.56} & \textbf{13.33} & \textbf{54.55} & 56.19 & \textbf{10.63} & 14.83 & \textbf{56.25} & \textbf{53.11} & \textbf{8.29} & 26.33 \\
\bottomrule
\end{tabular}
}
\end{table*}
\begin{table*}[t]

\begin{center}
\caption{Comparison on MathVista (Qwen2.5-VL-7B, $\rho=0.1$).}
\label{supp_tab:mathvista}
\small
\begin{tabular}{l|ccccccccc}
\toprule
\textbf{Method} & Full Cache & StreamLLM & H2O & SnapKV & LOOK-M \\ 
\textbf{MathVista} & 68.2 & 40.8 & 50.9 & 50.9 & 47.3 \\ \midrule
\textbf{Method} & MEDA & VLCache & Quest & PyramidKV & \textbf{FlashCache} \\ 
\textbf{MathVista} & 51.6 & 22.9 & 22.9 & 21.1 & \textbf{52.6} \\
\bottomrule
\end{tabular}
\end{center}

\end{table*}
\begin{table*}[t]
\begin{center}
\caption{Ablation Study on low-pass filter cutoff factor $\gamma$ on LLaVA-OneVision-8B.}
\label{supp_tab:ab_gamma}
\begin{tabular}{l|ccccccccc}
\toprule
\textbf{$\gamma$}               & 0.1   & 0.2   & 0.3   & 0.4   & 0.5   & 0.6   & 0.7   & 0.8 & 0.9 \\ 
\midrule
INIAH & \textbf{27.50} & 24.69 & 18.44  & 14.06 & 11.88  & 11.25 & 10.00 & 10.31   & 10.00   \\
GPR1200   & 12.33  & \textbf{12.83}  & 11.50 & 12.17  & 11.67 & 11.83 & 12.33 & 12.33   & 12.17  \\
\bottomrule
\end{tabular}
\end{center}
\end{table*}

\section{More Observations}
\label{sec:More Observation}
In this section, we further supplement our previous observations of low-frequency concentration in the KV matrices and outlier KV phenomena. Additional experiments across more models and datasets, along with visualizations, are conducted.

\subsection{Frequency-domain energy distribution of the KV matrices}
\begin{figure*}
    \centering
    \includegraphics[width=\textwidth]{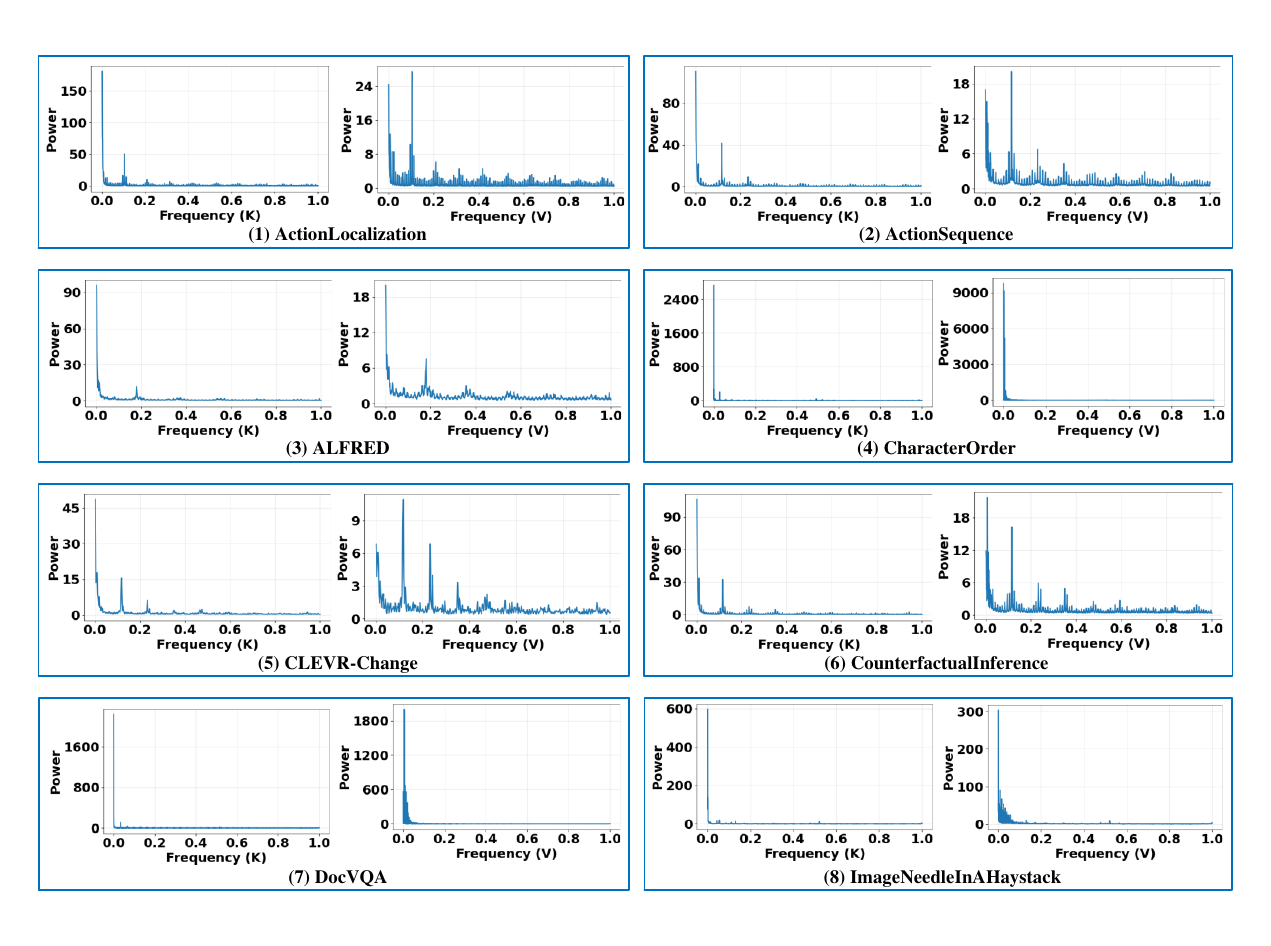}
    \caption{Frequency-domain energy distribution of the KV matrices.  Experiments are conducted on Qwen2.5-VL-7B-Instruct.}
    \label{supp_fig1}
\vspace{-3mm}
\end{figure*}

\begin{figure*}
    \centering
    \includegraphics[width=\textwidth]{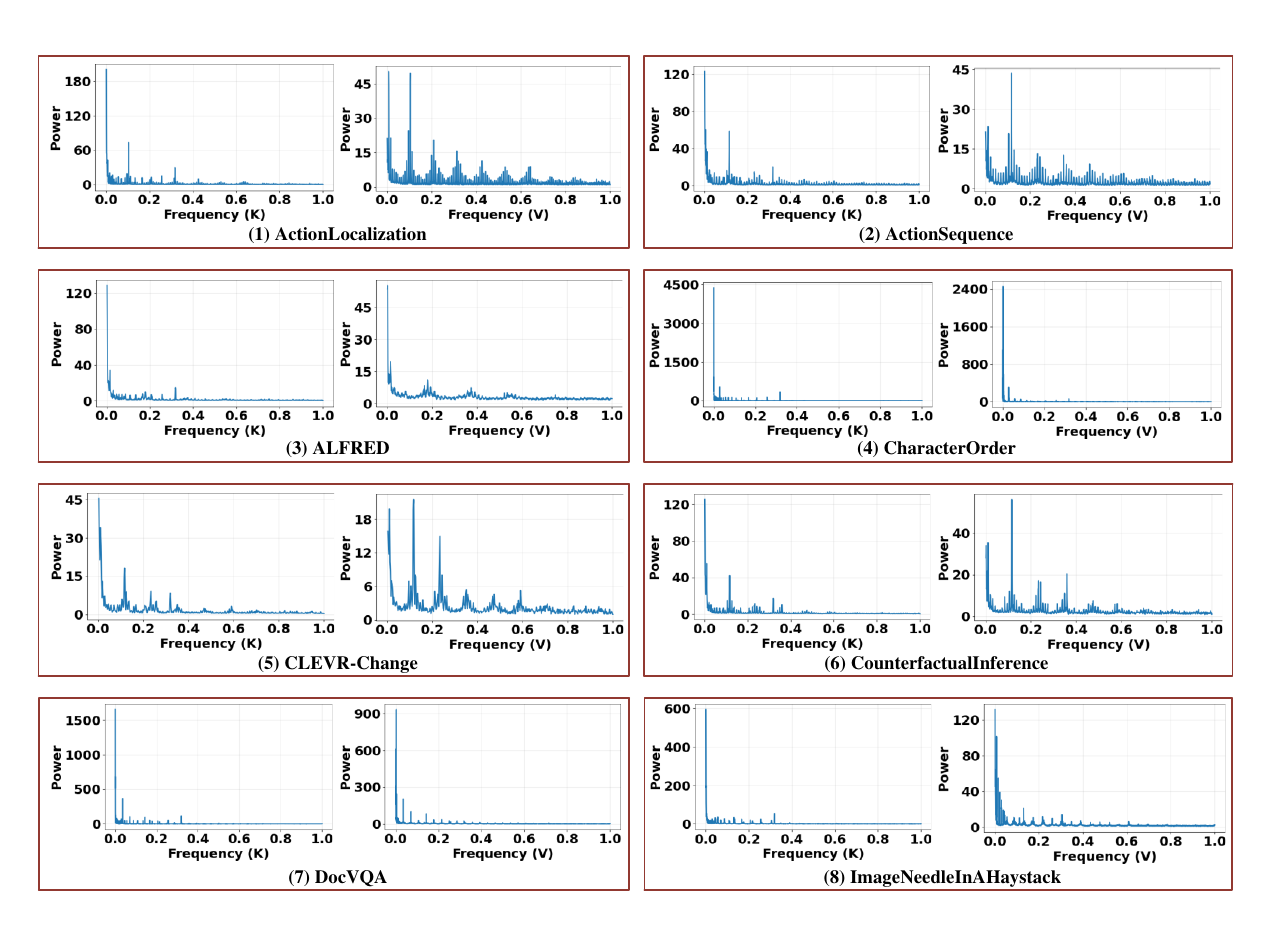}
    \caption{Frequency-domain energy distribution of the KV matrices. Experiments are conducted on LLaVA-OneVision-1.5-8B-Instruct.}
    \label{supp_fig2}
\vspace{-3mm}
\end{figure*}

We observe that the frequency-domain energy of KV matrices is predominantly concentrated at low frequency, with high frequency components occupying a relatively small proportion. To further validate this phenomenon, we conduct additional experiments across two models—Qwen2.5-VL-7B-Instruct~\cite{wang2024qwen2} and LLaVA-OneVision-1.5-8B-Instruct~\cite{LLaVAOneVision15} using eight datasets. Fig.~\ref{supp_fig1} presents the results for Qwen2.5-VL-7B-Instruct, while Fig.~\ref{supp_fig2} shows the results for LLaVA-OneVision-1.5-8B-Instruct. Combining the findings from both figures, we observe that the phenomenon of KV matrices frequency-domain energy concentration in the low-frequency range is significantly established.

\subsection{Frequency-domain energy distribution of KV matrices across different layers}
\begin{figure*}
    \centering
    \includegraphics[width=\textwidth]{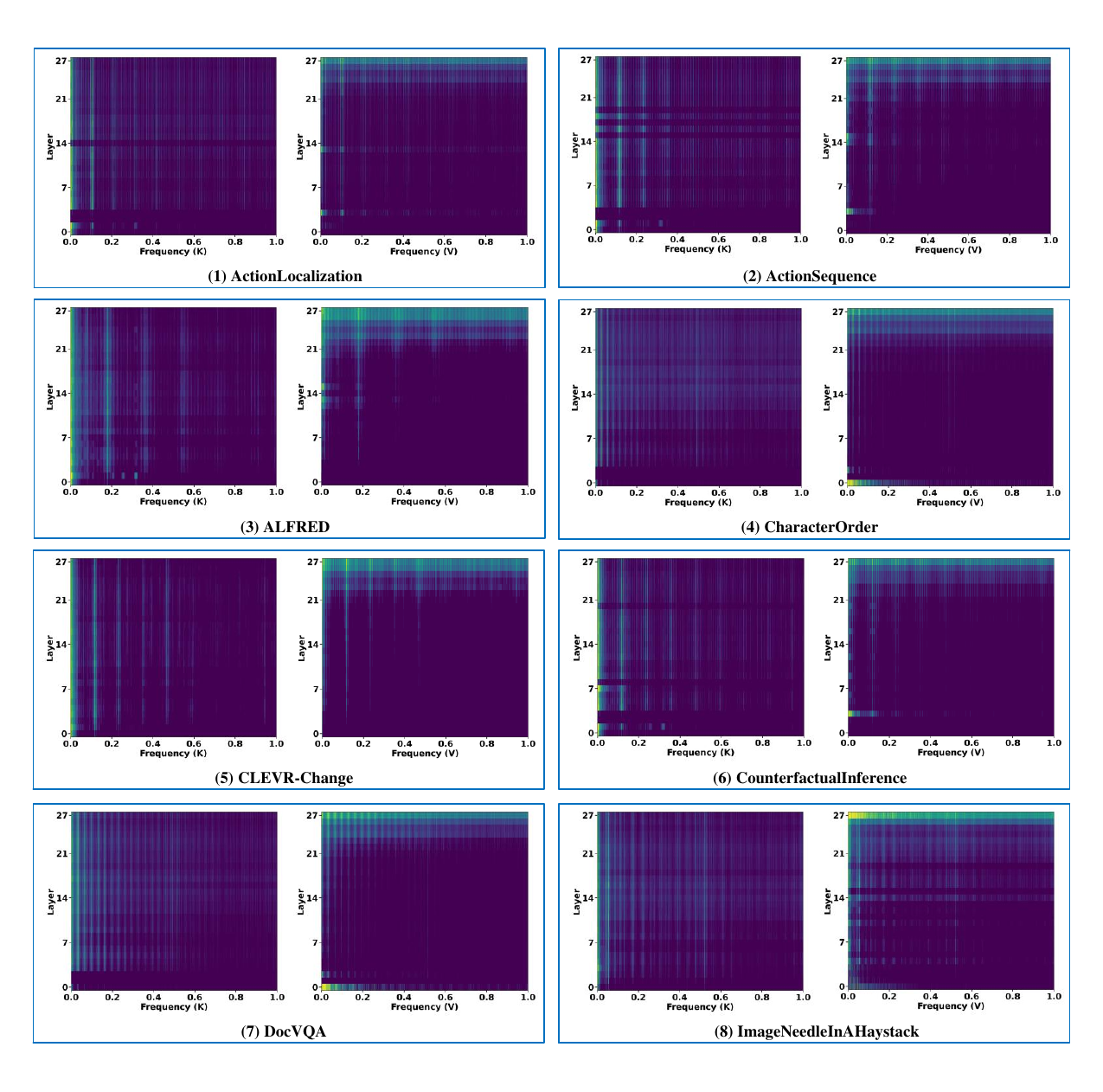}
    \caption{Frequency-domain energy distribution of KV matrices across different layers. Experiments are conducted on Qwen2.5-VL-7B-Instruct.}
    \label{supp_fig3}
\vspace{-3mm}
\end{figure*}

\begin{figure*}
    \centering
    \includegraphics[width=\textwidth]{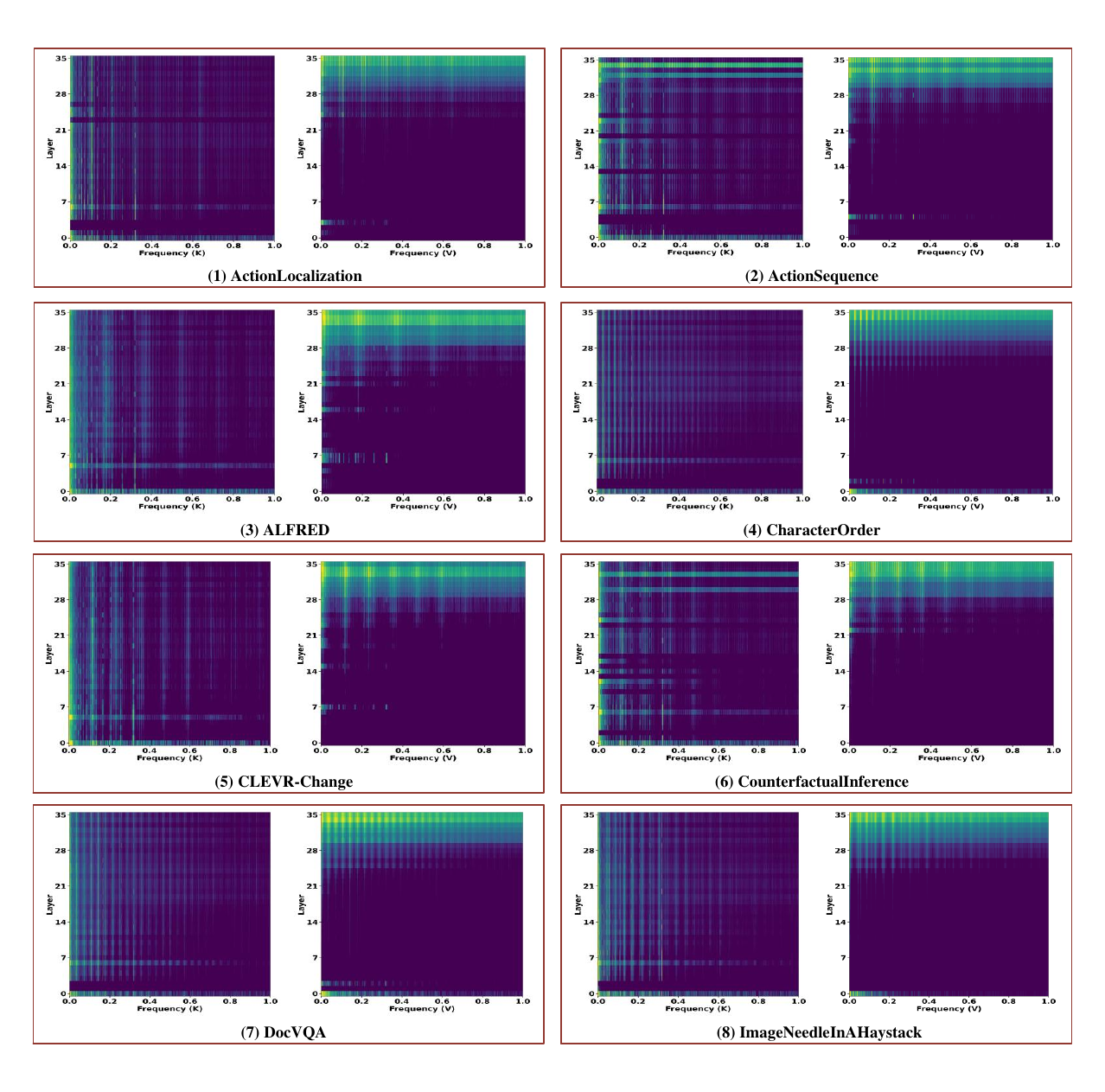}
    \caption{Frequency-domain energy distribution of KV matrices across different layers. Experiments are conducted on LLaVA-OneVision-1.5-8B-Instruct.}
    \label{supp_fig4}
\vspace{-3mm}
\end{figure*}

We observe clear differences in the frequency-domain energy distribution of KV matrices across layers. To further validate this phenomenon, we conduct additional experiments on two models—Qwen2.5-VL-7B-Instruct~\cite{wang2024qwen2} and LLaVA-OneVision-1.5-8B-Instruct~\cite{LLaVAOneVision15}—using eight datasets. The results for Qwen2.5-VL-7B-Instruct are shown in Fig.~\ref{supp_fig3}, and those for LLaVA-OneVision-1.5-8B-Instruct are presented in Fig.~\ref{supp_fig4}. Taken together, these results provide strong evidence that the frequency-domain energy distribution of KV matrices varies substantially across different layers.

\begin{table*}[t]
\begin{center}
\caption{Method Latency in per layer with 32K/64K sequence length (metric: ms).}

\label{supp_tab:Method Latency}
\begin{tabular}{c|cccccc}
\toprule
Method & Prefilling &H2O      & SnapKV  &LOOK-M   & MEDA  & FlashCache \\
\midrule
32K     &53.63& 1297.27   & 76.62   & 1290.64   & OOM  & \textbf{12.45}       \\
64K     &148.08& OOM & OOM   & OOM  & OOM & \textbf{28.12}       \\
\bottomrule
\end{tabular}
\end{center}
\end{table*}
\begin{table}[t]
\caption{More Efficiency Analysis}
\label{supp_tab:More Efficiency Analysis}
\resizebox{0.48\textwidth}{!}{
\begin{tabular}{ccccc}
\toprule
Metric & \multicolumn{2}{c}{Cache Size(GB)} & \multicolumn{2}{c}{Total Time(s)} \\
\midrule
Method & Full KV        & FlashCache        & Full KV        & FlashCache       \\
\midrule
2K     & 0.11           & \textbf{0.01}              & 19.93          & \textbf{19.54}            \\
4K     & 0.21           & \textbf{0.02}              & 20.53          & \textbf{19.56}            \\
8K     & 0.43           & \textbf{0.03}              & 22.12          & \textbf{19.62}            \\
16K    & 0.85           & \textbf{0.09}              & 24.57          & \textbf{20.28}            \\
32K    & 1.71           & \textbf{0.17}              & 30.51          & \textbf{21.75}            \\
64K    & 3.42           & \textbf{0.34}              & 43.5           & \textbf{25.31} \\      \bottomrule    
\end{tabular}
}
\end{table}

\subsection{The impact on model performance of prioritizing the removal of KV pairs under different removal ratios}
\begin{figure*}
\centering
\includegraphics[width=0.9\linewidth]{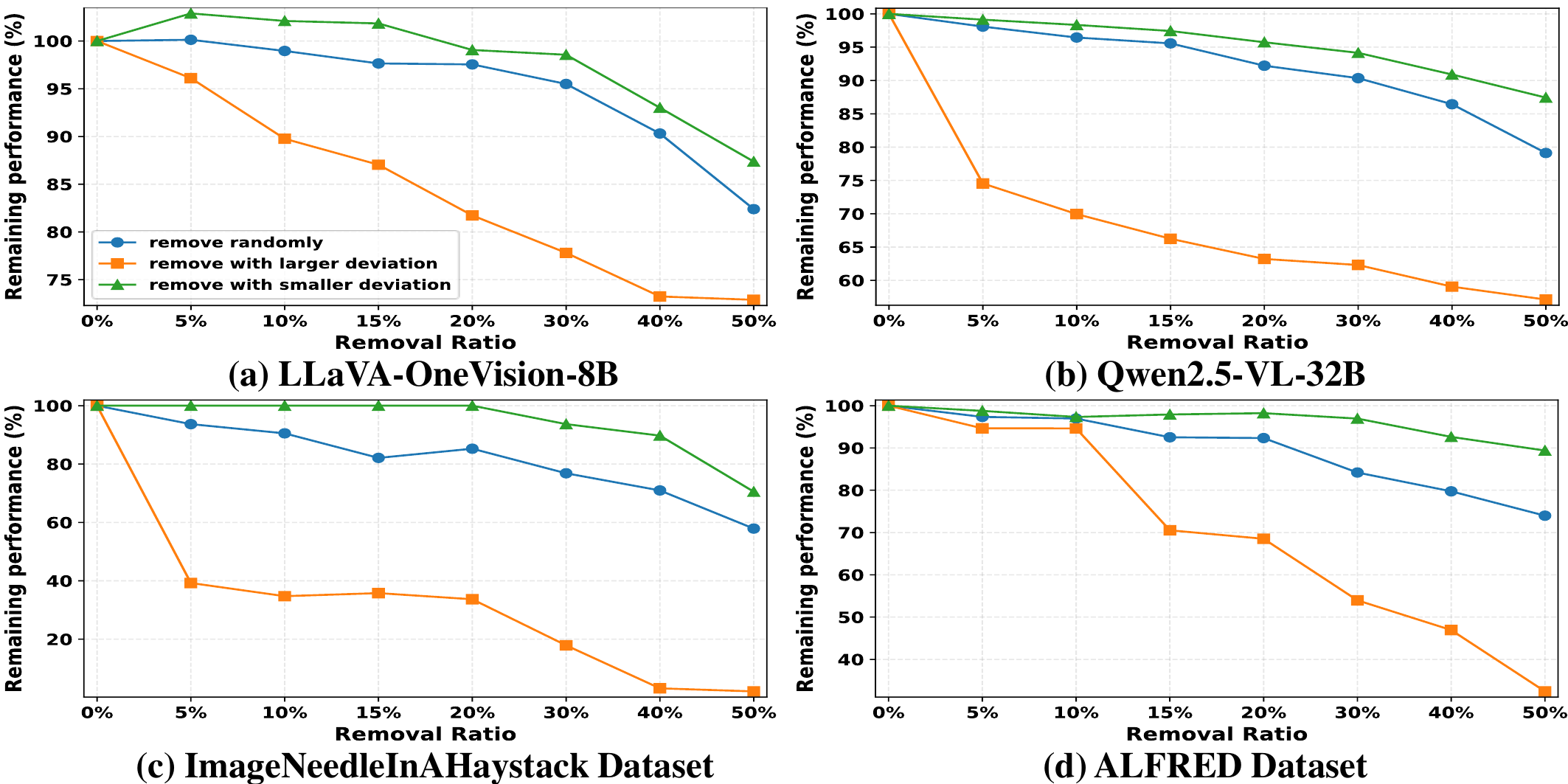}
\caption{More observation of "Outlier KV".}
\label{fig:outlierkv}
\vspace{-3mm}
\end{figure*}

In Fig.\ref{fig:outlierkv}, the frequency-domain outlier phenomenon is consistent across diverse model architectures, parameter scales, and datasets. This evidence substantiates that our core insight is inherently general rather than model/dataset-specific.

\section{More Efficiency Analysis}
\label{sec:More Efficiency Analysis}

In this section, we further analyze the efficiency of FlashCache. Tab.~\ref{supp_tab:More Efficiency Analysis} reports the actual KV cache memory footprint and end-to-end inference time as the input length varies, under a 10\% KV retention ratio. We measure the memory footprint and the end-to-end inference time for generating 512-token outputs to obtain per-token latency across input lengths {2K, 4K, 8K, 16K, 32K, 64K}. All experiments are conducted with FlashAttention on Qwen2.5-VL-7B-Instruct. The results show that FlashCache substantially reduces both the KV cache memory footprint and the end-to-end inference time, with the acceleration effect becoming increasingly pronounced as the input length grows.

As shown in Tab.\ref{supp_tab:Method Latency}, \textit{FlashCache}'s overhead is negligible (12.45ms at 32K). Unlike baselines that suffer from OOM due to attention re-computation, \textit{FlashCache} eliminates attention dependency, ensuring memory stability at 64K while introducing substantially lower prefilling overhead than existing methods.

\section{Case Study}
\label{sec:Case Study}
\begin{figure*}
    \centering
    \includegraphics[width=0.6\textwidth]{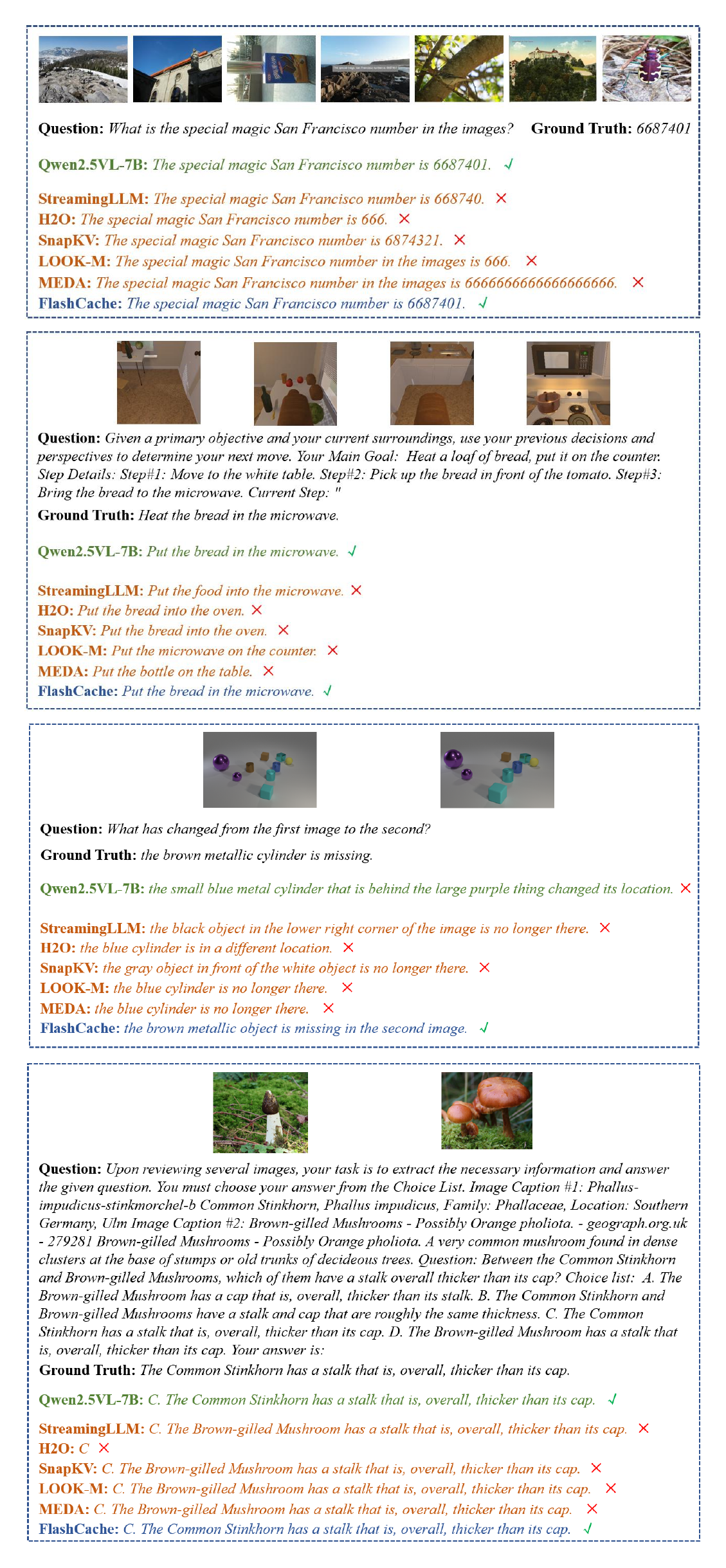}
    \vspace{-3mm}
    \caption{Case Study}
    \label{supp_fig5}
\end{figure*}

To further demonstrate the advantages of our approach, we have selected several reasoning examples, as shown in Fig.~\ref{supp_fig5}. We compare FlashCache against state-of-the-art KV cache eviction methods:
StreamingLLM~\cite{xiao2023efficient}, H2O~\cite{zhang2023h2o}, SnapKV~\cite{li2024snapkv}, LOOK-M~\cite{wan2024look}, and MEDA~\cite{wan2025meda}. We conduct comparisons across different image lengths, different text lengths, and different tasks. KV Cache retention ratio is setting as 0.05. Experimental results indicate that under high compression ratios, FlashCache preserves more of the model's inherent performance characteristics. At the same time, we find that FlashCache enables the model to answer questions it previously could not. We believe FlashCache filters out noise in some input data, thereby improving the model's performance under certain conditions.

\section{Benchmarks}
\label{sec:Benchmarks}

\subsection{Multi-Images Understanding Benchmarks}
\textbf{MileBench\cite{song2024milebench}.}~MileBench is a multimodal long-context benchmark designed to evaluate how well Multimodal Large Language Models handle complex multi-image inputs over extended contexts. It is built from 6,440 long-context samples collected from 29 publicly available or self-constructed datasets, with each sample containing multiple images and accompanying text. In our paper, we use four task types from MileBench: Task T (Temporal Multi-Image), which focuses on understanding and reasoning over sequences of images with temporal relationships; Task S (Semantic Multi-Image), which requires integrating complementary semantic information distributed across multiple images; NH (Needle in a Haystack), which tests the model’s ability to retrieve a small but crucial visual or textual detail from long, cluttered multimodal contexts; and IR (Image Retrieval), which assesses whether the model can accurately locate or identify the correct image given a textual description or multimodal query. Together, these tasks make MileBench a comprehensive tool for probing long-context multi-image comprehension, reasoning, and retrieval abilities.

\textbf{MUIRBench\cite{wang2024muirbench}.}~MUIRBench is a comprehensive benchmark for robust multi-image understanding in multimodal large language models. It contains 11,264 images and 2,600 multiple-choice questions, covering 12 diverse multi-image understanding tasks (such as scene understanding, temporal ordering, and visual retrieval) and 10 categories of inter-image relations (including multiview, narrative, and complementary relations). Each example is constructed in a pairwise manner, where an answerable instance is paired with a minimally modified unanswerable variant, enabling robust evaluation of whether models truly integrate information across multiple images rather than relying on superficial cues. This design makes MUIRBench a strong stress test for multi-image reasoning and a useful complement to standard single-image vision–language benchmarks.

\textbf{MMMU\cite{yue2023mmmu}.}~MMMU (Massive Multi-discipline Multimodal Understanding) is a large-scale benchmark designed to evaluate foundation models on expert-level, university and beyond problems across a wide range of real-world disciplines. It contains over 10,000 multimodal multiple-choice questions spanning fields such as medicine, law, physics, chemistry, engineering, and the humanities, with inputs that combine text, diagrams, charts, tables, and other images. Many questions involve not just a single image but multiple related figures or panels (for example several diagrams, tables, or subplots for one problem) that must be interpreted jointly, allowing the benchmark to test genuine multi-image reasoning rather than isolated image understanding. All items are carefully curated from real exam or professional materials, emphasizing high-level reasoning, domain knowledge, and cross-modal comprehension, and are widely used to assess how close multimodal large language models are to human expert performance in complex, knowledge-intensive scenarios.

\subsection{High Resolution Benchmarks}
\textbf{V*\cite{vstar}.}~V* is an LLM-guided visual search mechanism designed to help multimodal large language models actively locate task-relevant regions in an image instead of passively encoding the whole frame. When paired with an MLLM, it underpins the SEAL (“Show, Search, and Tell”) meta-architecture, which iteratively queries the image to refine visual grounding and reasoning. To evaluate this capability, the authors introduce V* Bench, a dedicated benchmark built from 191 high-resolution natural images (average resolution around 2246×1582) that emphasize fine-grained details and visually crowded scenes. V* Bench focuses on detailed visual grounding through attribute recognition and spatial relationship reasoning questions, making it particularly suitable for testing whether models can reliably handle high-resolution inputs and answer questions about small or easily overlooked visual elements.

\textbf{HR-Bench\cite{hrbench}.}~HR-Bench is a high-resolution visual question answering benchmark designed to assess the fine-grained perception ability of multimodal large language models on ultra high-resolution images. It consists of two sub-tasks: Fine-grained Single-instance Perception (FSP), which contains 100 samples focusing on detailed attribute recognition, OCR, and visually grounded prompting within a single image, and Fine-grained Cross-instance Perception (FCP), which contains 100 samples targeting map understanding, chart analysis, and spatial relationship reasoning across multiple visual elements. HR-Bench is provided in both 8K and 4K settings: the 8K version uses images with approximately 8K-level resolution, while the 4K version is obtained by carefully annotating objects in the 8K images and cropping them into 4K regions, enabling rigorous evaluation of whether models can maintain and exploit fine details under realistic high-resolution conditions.

\subsection{Video Understanding Benchmark}
\textbf{FAVOR-Bench\cite{tu2025favor}.}~FAVOR-Bench is a comprehensive benchmark for fine-grained video motion understanding that targets the ability of multimodal large language models to perceive and reason about detailed temporal dynamics rather than just coarse scene changes. It contains 1,776 videos from both ego-centric and third-person perspectives, each accompanied by structured manual motion annotations. On top of this data, FAVOR-Bench provides both close-ended and open-ended evaluation settings: for close-ended evaluation, it offers 8,184 multiple-choice question–answer pairs organized into six motion-centric sub-tasks, while for open-ended evaluation it includes a GPT-assisted caption assessment protocol together with a cost-efficient LLM-free evaluation method. Through these tasks, FAVOR-Bench enables systematic measurement of how well models recognize, localize, and describe fine-grained motions in realistic videos, and serves as a challenging testbed for improving motion-sensitive video understanding.

\section{Models}
\label{sec: Models}
\subsection{Qwen2.5-VL-7B-Instruct}
Qwen2.5-VL-7B-Instruct~\cite{wang2024qwen2} is a 7-billion-parameter vision–language model from the Qwen2.5-VL family, instruction-tuned to handle a wide range of multimodal tasks with text, images, documents, charts, and video as input. It combines a strong language backbone with a visual encoder that supports dynamic resolutions and long-context processing, enabling capabilities such as document analysis, OCR, chart and layout understanding, temporal reasoning over videos, and structured output generation (for example JSON and bounding boxes for visual localization).

\subsection{LLaVA-OneVision-1.5-8B-Instruct}
LLaVA-OneVision-1.5-8B-Instruct~\cite{LLaVAOneVision15} is an 8-billion-parameter large multimodal model in the LLaVA-OneVision-1.5 family, designed as a fully open, efficient framework for high-quality multimodal training. It is trained on native-resolution images and curated instruction data, aiming to deliver strong performance on diverse vision–language benchmarks while keeping computational costs relatively low. The model follows the LLaVA-style visual instruction tuning paradigm and is positioned as a general-purpose multimodal assistant capable of fine-grained image understanding, reasoning, and dialogue under an open-source training and evaluation pipeline.

\section{Limitation and Future Work}
In the future, we will continue to explore efficient compression for multimodal KV Cache without attention scores, striving to extend its application to more scenarios such as embodied intelligence and ultra-long context.

\newpage

\end{document}